\documentclass[letterpaper, 10 pt, conference]{ieeeconf}  

\IEEEoverridecommandlockouts                              
\usepackage[utf8]{inputenc}
\usepackage[T1]{fontenc}
\usepackage{multirow}
\usepackage{graphicx}
\usepackage{svg}
\usepackage{amsmath}
\usepackage{amsfonts,amssymb}
\usepackage{xcolor}

\title{\LARGE \bf
Data-driven Holistic Framework for Automated Laparoscope Optimal View Control with Learning-based Depth Perception
}

\author{Bin Li$^{1}$, Bo Lu$^{1,*}$, Yiang Lu$^{1}$, Qi Dou$^{2}, \textit{Member, IEEE}$, and Yun-Hui Liu$^{1}, \textit{Fellow, IEEE}$
\thanks{This work is supported in part of the HK RGC under T42-409/18-R and 14202918, in part by the Shenzhen-HK Collaborative Development Zone, in part by the Multi-Scale Medical Robotics Centre, InnoHk, and in part by the VC Fund 4930745 of the T Stone Robotics Institute.}
\thanks{$^{1}$B. Li, B. Lu, Y. A. Lu, and Y. H. Liu are with the T stone Robotics Institute, The Department of Mechanical and Automation Engineering, The Chinese University of Hong Kong. Email: bolu@cuhk.edu.hk.
*Corresponding author.}%
\thanks{$^{2}$Q. Dou is with the Department of Computer Science and Engineering, and T Stone Robotics Institute, The Chinese University of Hong Kong.}%
}

\begin{document}

\maketitle
\thispagestyle{empty}
\pagestyle{empty}

\begin{abstract}

Laparoscopic Field of View (FOV) control is one of the most fundamental and important components in Minimally Invasive Surgery (MIS), nevertheless the traditional manual holding paradigm may easily bring fatigue to surgical assistants, and misunderstanding between surgeons also hinders assistants to provide a high-quality FOV.
Targeting this problem, we here present a data-driven framework to realize an automated laparoscopic optimal FOV control. To achieve this goal, we offline learn a motion strategy of laparoscope relative to the surgeon’s hand-held surgical tool from our in-house surgical videos, developing our control domain knowledge and an optimal view generator.
To adjust the laparoscope online, we first adopt a learning-based method to segment the two-dimensional (2D) position of the surgical tool, and further leverage this outcome to obtain its scale-aware depth from dense depth estimation results calculated by our novel unsupervised RoboDepth model only with the monocular camera feedback, hence in return fusing the above real-time 3D position into our control loop.
To eliminate the misorientation of FOV caused by Remote Center of Motion (RCM) constraints when moving the laparoscope, we propose a novel rotation constraint using an affine map to minimize the visual warping problem, and a null-space controller is also embedded into the framework to optimize all types of errors in a unified and decoupled manner.
Experiments are conducted using Universal Robot (UR) and Karl Storz Laparoscope/Instruments, which prove the feasibility of our domain knowledge and learning enabled framework for automated camera control.

\end{abstract}

\section{INTRODUCTION}
Minimally Invasive Surgery (MIS) is of great significance because it has benefits of less trauma, less pain and faster recovery compared with open surgery. In conventional MIS, a surgical assistant is required to manipulate the laparoscope to provide the surgeon with a proper view for operation. Hence, designing robots, e.g., EndoAssist \cite{aiono2002controlled}, ViKY EP, AESOP, to manipulate the laparoscope is a popular trend. These on-the-shelf systems are manipulated via the direct hand operation, foot pedal, eye tracking \cite{fujii2018gaze}\cite{gras2017implicit}, etc. Nevertheless, these control signals sent by surgeons may distract their attentions and increase operational burdens. To further liberate surgeons from tedious tasks, automation of robot-assisted laparoscopic Filed of View (FOV) control is crucial.

\begin{figure}[tpb]
      \centering
      \includegraphics[width = 1.0\hsize]{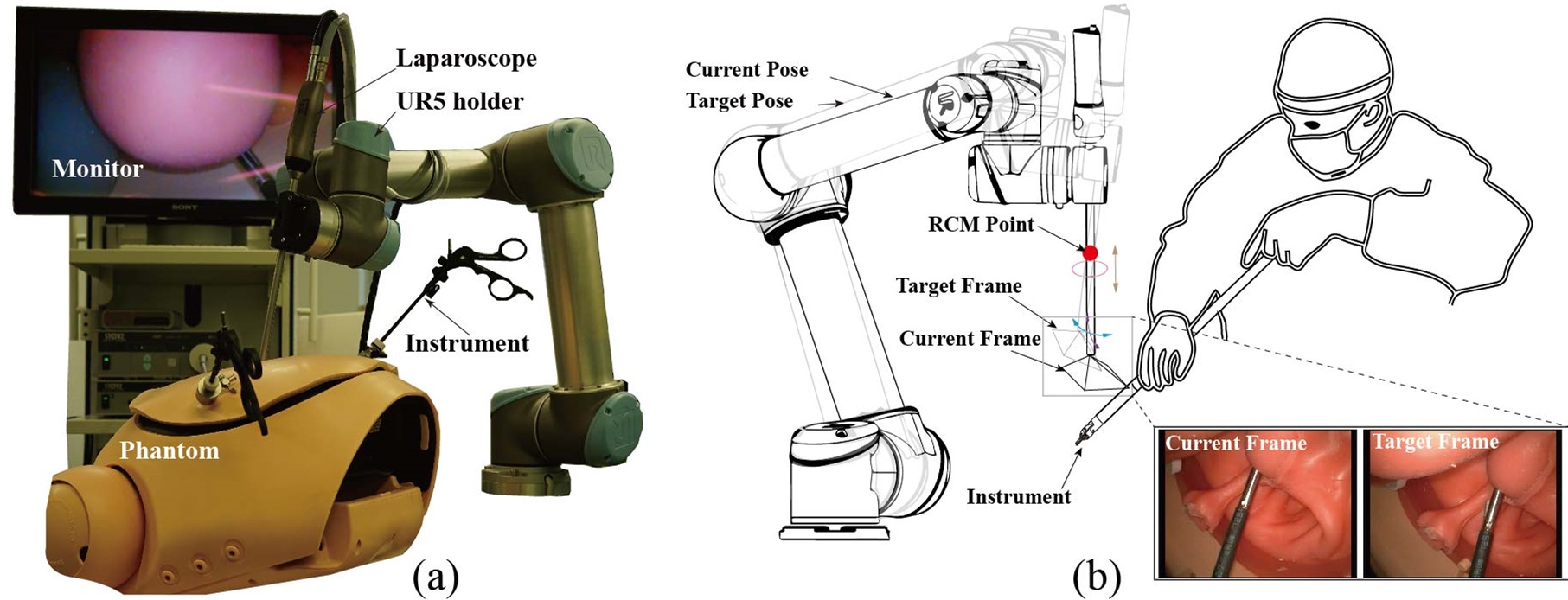}
      \caption{
      Illustration of an automated FOV control of laparoscope. (a). Robotic and laparosocopic system; (b). Optimal FOV planning for surgeons.
      }
      \label{System Setup}
    \vspace{-0.55cm}
   \end{figure}

During the past decade, lots of researchers have put their attention on this task. Osa \textit{et al.} \cite{osa2010framework} proposed a visual servoing method which moves the laparoscope automatically to keep the tool at the center of FOV. Yang \textit{et al.} \cite{yang2019adaptive} further introduced a method to realize a FOV control by placing the tool within an arbitrary yet well-defined closed region. However, their strategies are rigidly pre-programmed and lack the intelligence.
Facing this problem, Bihlmaier \textit{et al.} \cite{bihlmaier2014automated} proposed a cognitive endoscope robot by training a camera quality classifier with expert annotations and predefined guidance strategy to choose the next optimal viewpoint. Agrawal \cite{agrawal2018automating} presented an imitation learning-based method to generate the optimal endoscopic movement according to the current phase. However, these methods require expensive data collections and annotations, which are inconvenient to be generalized and transferred to new systems.
To resolve this problem, we propose to use large amounts of unlabeled legacy surgical videos and extract a domain knowledge for an intelligent FOV control. (Fig. \ref{System Setup})

Apart from the domain knowledge, the real-time 3D position of the main surgical tool is highly demanded when automating the laparoscopic FOV control. Considering there are no stable features on the smooth and reflective surface of the metal tool, some researchers adopted colored \cite{osa2010framework}\cite{king2013towards} or artificial markers \cite{zhang2017real} to first locate its image position, and indirectly estimate the depth by calculating the marker's relative area in the image.
Doignon \textit{et al.} proposed a cylinder pose estimation method \cite{doignon2007degenerate} that can be further utilized for tool tip depth calculation, while its heavy computational cost cannot be deployed for real-time operations. 
To overcome the shortcomings in traditional methods, learning-based approaches, especially the unsupervised learning ones \cite{zhou2017unsupervised}\cite{godard2019digging}\cite{gur2019single}\cite{lu2020multi} which are independent of expensive labeling data, are widely adopted recently. However, based on monocular laparoscopic feedback, such unsupervised methods can only recover a relative depth. To obtain the scale-aware depth, prior knowledge, such as the constant camera height assumption \cite{zhou2019ground}\cite{song2015high}\cite{wagstaff2020self} or the object size \cite{frost2018recovering}\cite{frost2016object}, is needed. To the best of our knowledge, the scale-aware depth estimation for surgical tool using a monocular unsupervised learning method has not been exploited. 

In addition, another so-called surgical view misorientation problem \cite{holden1999perceptual} may frequently happen, in which the line-of-sight from the camera can be different from the surgeon's desired view when he/she looks directly into the abdomen.
Such problem can be even severer when the camera moves under Remote Center of Motion (RCM) constraint.
Conceptual solutions \cite{breedveld1999theoretical}\cite{wentink2000endoscopic} have been studied and depend on manually manipulation. 
Kurt \textit{et al.} presented a digitally rotation correction method by using an extra Inertial Measurement Unit (IMU) to detect the orientation \cite{holler2009endoscopic}. However, it relies on external sensors and does not study the control issues of endoscope.
Our previous work \cite{yang2019adaptive} also tackled the misorientation problem in a subjective manner. However, it lacks physical meaning and still encounters the misorientation problem when the camera tracks the surgical tool with large motions.

Regarding all these challenges, we propose a data-driven framework to automate the laparoscopic optimal FOV control. Our main contributions can be summarized as follows:
\begin{itemize}

\item We propose a novel monocular unsupervised learning method named RoboDepth to achieve an accurate scale-aware depth estimation for surgical tools. This method can obtain real-time computing and allow simultaneous motions of the tool and laparoscope.
\item By analyzing from large unlabeled surgical videos, we empirically generate a domain knowledge, which can be adopted to formulate an optimal view generator to perform FOV control in an expert-like manner.
\item We introduce a novel affine mapping-based Minimize Rotation Constraint (MRC) method, which is enabled with distinct physical meaning to solve the visual misorientation problem. 
\item Integrating the above enhancements, we design a novel null-space controller to automate the motion of the robot-held laparoscope. It maintains proper 2D and 3D positions of the main tool w.r.t. the laparoscope, while eliminating the misorientation error in a unified way.



\end{itemize}

\section{METHODOLOGY}

   \begin{figure*}[thpb]
      \centering
      \includegraphics[width = 1.0\hsize]{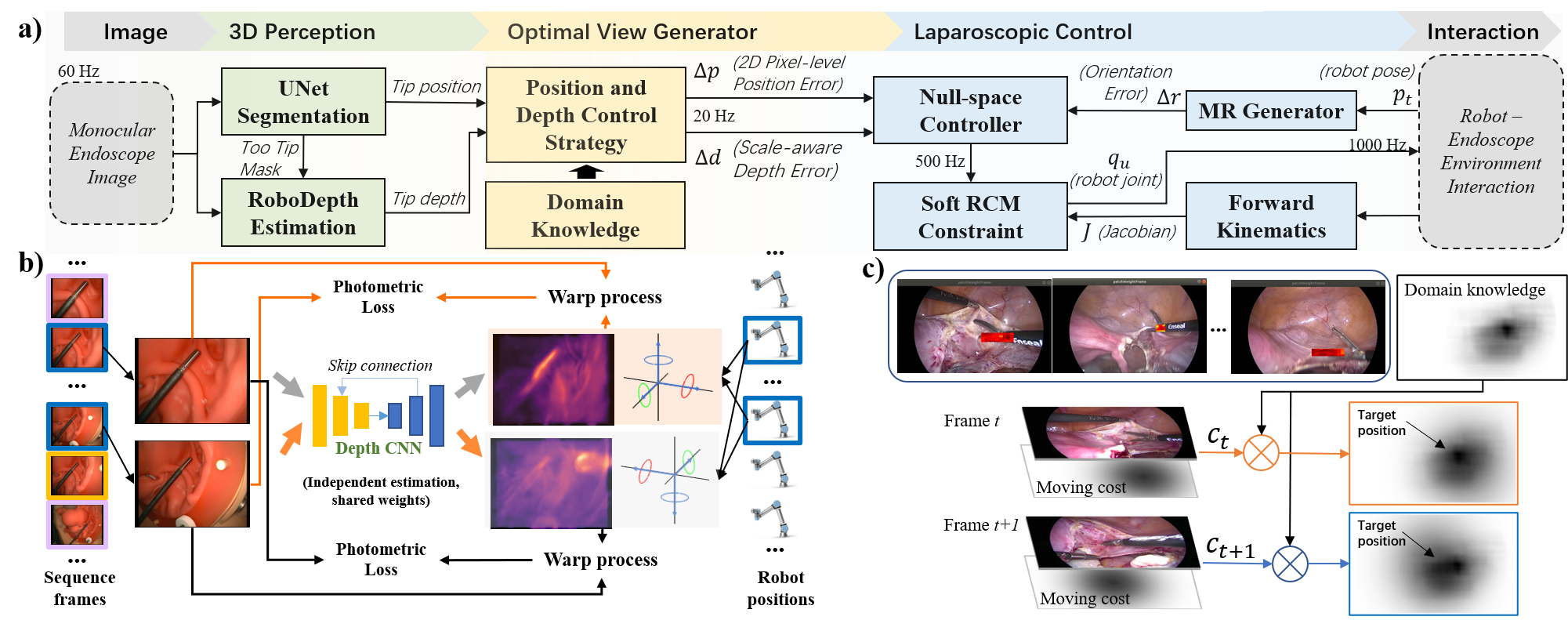}
      \caption{(a). Workflow of the automated FOV control; (b). The training workflow of RoboDepth model for instrument depth estimation; (c). The frequency heatmap that obtained by statistics from surgical video data can be served as the domain knowledge for a laparoscope 2D control.}
      \label{overall_Framework}
      \vspace{-0.50cm}
   \end{figure*}

The overview of our proposed framework is shown in Fig. \ref{overall_Framework}. By modifying a U-Net based segmentation model, a marker-free method is used to obtain the tool's pixel-wise position. Leveraging the scale-aware depth estimated by RobotDepth, its 3D coordinates w.r.t. the laparoscope can be achieved. Adopting the data-driven domain knowledge, we design an optimal view generator by balancing the weight between the moving cost and the view's optimality. Combining the online perception and offline domain knowledge together, we hence formulate a novel null-space controller to automate the laparoscopic motion, while eliminating the visual misorientation under RCM motion constraint. The details of each function block are presented in the following.

\subsection{Perception of 3D Position of Surgical Instrument}

\subsubsection{Instrument Tip Segmentation Model}

To initialize the task, a real-time 2D segmentation of the tool tip is essential. 
Owing to its exquisite architecture, U-Net \cite{ronneberger2015u} that possesses a fast processing time can output accurate segmentation results with a small amount of training data. Hence, we adopt this architecture with VGG11 as the backbone to train our tip segmentation model.
To ensure its generality, training laparoscopic images are collected with multiple tip conditions, such as various jaw angles, tool poses, and etc. 
To further enhance the segmentation performance, a multi-task loss function is designed:
\begin{equation}
Loss = w_{1} L_\text{Cross Entropy}+w_2 L_\text{Lovasz}+w_3 L_\text{Generalized Dice}
\end{equation}
where $w_1=0.1$, $w_2=0.1$, and $w_3=0.8$ are weighting parameters of three loss components that jointly highlight metal features of the tool tip.
Then we obtain the tool tip position $p_t$ by calculating the center of mass of the segmented binary mask.

 


\subsubsection{Scale-aware Depth Estimation Model}
To recover the scale-aware depth $d_{tool}$ from monocular images, extra physical constraints, e.g. the diameter of the instrument, are needed. To fully utilize the robotic information, we present a novel monocular depth estimation network named RobotDepth that can use a single image input $I_t$ to produce a dense scale-aware depth map $D_t$. Our core idea is to merge our robotic kinematics into the neural network as physical constraints to recover an accurate depth. To the best of our knowledge, this is the first work integrating such robotic knowledge to the learning architecture to estimate a scale-aware depth.

\textbf{Data Collection:}
Attaching the laparoscope at the end-effector of UR5 robot, we can collect the monocular image and calculate its corresponding six-dimensional (6D) pose $^b{\textbf{T}}{_c}$ by utilizing the pre-calibrated transformation from end-effector to laparoscope $^e{\textbf{T}}{_c}$ and the end-effector pose $^b{\textbf{T}}{_e}$:
$^b{\textbf{T}}{_c} = {}^b{\textbf{T}}{_e}\cdot {}^e{\textbf{T}}{_c}$.
To diversify the collected samples, we collect 30 sequences of separate videos and their corresponding 6D pose by placing instruments and surgical phantoms at varying conditions. In each video, the surgical tool maintains stable while the laparoscope is moved with 1 $mm$ step discrepancy to record around 120 samples.

\textbf{Sampling Strategy:}
Since the difference between successive 6D pose in each sequence is very small, a hierarchical scheme is adopted to sample the training data.
The sequence size is $N$ and total number of $\log_2(N-1)$ levels will be sampled in each sequence. The level $l$ contains a continuous and sparse sampling of frames $I_{m(n)}$: $S_{l}=\left\{(m, n)| \mid m-n\mid=2^{l}, m \bmod 2^{l-1}=0\right\}$ and the final sampling frame set in all levels is $S=\bigcup_{0 \leq l \leq\left\lfloor\log _{2}(N-1)\right\rfloor} S_{l}$. During the training process, each sampled image pair $I_{m}$ and $I_n$ based on the above hierarchical scheme is served as the training data.

\textbf{Loss Formulation:}
Given a sequence of monocular images and corresponding 6D poses, we formulate an unsupervised learning framework as shown in Fig. \ref{overall_Framework}(b). 
For Image $I_m$ and $I_n$, the depth CNNs, which utilizes the encoder and decoder structure \cite{godard2017unsupervised}, can estimate their corresponding dimensionless disparity map $Disp_{m(n)} \in [0, 1]$. Here we design a function $f$ to convert the disparity to the depth $D_{m(n)}$, which is expressed as:
\begin{equation}
\resizebox{0.90\hsize}{!}{$
D_{m(n)} = f\left(Disp_{m(n)}\right) = \frac{D_{min}D_{max}} {D_{min} + \left(D_{max}-D_{min}\right) \cdot Disp_{m(n)}}$}
\end{equation}
where $D_{min}$ and $D_{max}$ are chosen as 1 $mm$ and 100 $mm$ respectively to constrain the range of our estimated depth map $D_{m(n)}$. The median depth in the segmented tool area from $D$ is taken as the tool tip depth $d_{tool}$.
Let $p^{ij}_n\in \mathbb{R}^{3\times 1}$ denotes the homogeneous coordinates of a pixel in image $I_n$ and $K\in \mathbb{R}^{3\times 3}$ represents the camera intrinsic matrix, we can use the camera projection model to recover 3D coordinates of $p^{ij}_n$ and utilize the rigid transformation relationship ${}^b\textbf{T}_m^{-1}\cdot {}^b\textbf{T}_n$ recorded from robot to warp this 3D position to its corresponding point when the laparoscope captures image $I_m$. By projecting it to the image $I_m$, we can hence get the estimated corresponding 2D position of $p^{ij}_n$ in image $I_m$, which can be written as:
\begin{equation}
p_{m^{\prime}}^{i^{\prime} j^{\prime}} \sim K \cdot \underbrace{{}^b{\textbf{T}}^{-1}_{m}\cdot ^b{\textbf{T}}{_n}}_{pose \ n \ to \ m} \cdot \underbrace{D_{n}^{ij} \cdot K^{-1} p_{n}^{ij}}_{3D \ point}
\label{warping}
\end{equation}

Similarly, the estimated $p_{n^{\prime}}^{i^{\prime} j^{\prime}}$ can also be derived. By applying this calculation to all pixels in the image, the warped images $I_m^{\prime}$ and $I_n^{\prime}$ can be achieved. When the depth estimation is accurate, there should not exist different between $I_{m(n)}$ and $I_{m(n)}^{\prime}$. According to this principle, our training network is formulated which requires to minimize the reconstruction (Siamese) loss $\mathcal{L}_{re}$:
\begin{equation}
\mathcal{L}_{re}=\sum_{m,n} \text{Loss}\left(I_{m}, I_{m}^{\prime}\right)+\text{Loss}\left(I_{n}, I_{n}^{\prime}\right)
\end{equation}

In the above equation, Loss$\left(\cdot\right)$ is defined as:
\begin{equation}
\resizebox{0.90\hsize}{!}{
$\text{Loss}\left(I_{k}, I_{k}^{\prime}\right)=\frac{\alpha}{2}\left(1-\text{SSIM}\left(I_{k}, I_{k}^{\prime}\right)\right)+(1-\alpha)\left\|I_{k}-I_{k}^{\prime}\right\|_{1}$}\end{equation}
where $I_{k}$ and $ I_{k}^{\prime}$ represents the original image and warped image respectively ($k=m \text{ or } n$), and SSIM$\left(\cdot\right)$ represents their similarity. It should be noticed we choose the average pool layer with size 3 instead of Gaussian operation to calculate SSIM$\left(\cdot\right)$ with the weight parameter $\alpha = 0.85$.

Based on the above work, we propose a novel RoboDepth model to perform depth estimation on both sampled $I_m$ and $I_n$ images using the same depth CNN model with shared weights during the training. This Siamese shape architecture enables us to fully utilize the training dataset and generalize the RoboDepth model to depth estimation at various conditions. 
To guarantee the smoothness of the depth map, we integrate an edge-aware smoothness loss
$\mathcal{L}_{s,k}$ to discourage the shrinking, which is:
\begin{equation}
    \mathcal{L}_{final}=\sum_{l}\mu \mathcal{L}_{re}^{l}+\lambda \mathcal{L}_{s, k}^{l}
\end{equation}
where $\mu=0.8$ and $\lambda=0.2$ are weighs for each individual loss, $l$ denotes the loss calculation from different image scales \{$1$, $\frac{1}{2}$, $\frac{1}{4}$, $\frac{1}{8}$\}, $\mathcal{L}_{s,k}=\left|\partial_{x} D_{k}\right| e^{-\left|\partial_{x} I_{k}\right|}+\left|\partial_{y} D_{k}\right| e^{-\left|\partial_{y} I_{k}\right|}$, and $\partial_x$ and $\partial_y$ represent the partial derivatives along X and Y directions in the image level.


\subsection{Optimal View Generator}

\subsubsection{Domain knowledge}

To automate the laparoscope control, a proper moving policy is demanded. Traditional methods \cite{osa2010framework}\cite{yang2019adaptive} using hard-coding rules are straightforward yet lack surgical proficiency, which cannot be generalized in operating theaters. To overcome this problem, we turn to seek for domain knowledge that can guide the camera's motion and place the instrument tip in a surgeon-preferred area.  

Based on our in-house surgical data of hysterectomy, we statistically learn a data-driven optimal 2D image region in which surgeons prefer to place the instruments in one particular surgical phase (i.e. ligament dissection). To ensure the generality of this domain knowledge, we adopt PAWSS \cite{du2019patch}, as shown in Fig. \ref{overall_Framework}(c), to track the dynamic motion of the surgical tool in 3506 image frames, and we use statistic fitting to output an empirically reliable frequency heat map, in which a higher value represents a better place to put the surgical instrument during manipulation. 
It should be noticed the generation of such domain knowledge can be easily extended to other surgical phases, e.g. dividing peritoneum, dividing uterus vessels, etc.

\subsubsection{Position and Depth Control Strategy}

With the scale-aware depth of the instrument tip and its optimal 2D placing area in FOV, a control strategy for the laparoscopic motion is needed. 
To mimic its manipulation in practice, the strategy should jointly considering two criteria:\\
\textbf{Criteria 1:} The laparoscopic FOV should track the dynamic tool and put its tip within the optimal area, while avoiding frequent motions to reduce visual fatigue.\\
\textbf{Criteria 2:} When adjusting the camera, the strategy should balance its moving distance and the optimality of the 2D position for placing the tip.

Towards these targets, we dynamically calculate the raw image distances between the current 2D position of the tip and all pixels within the domain knowledge-based region, and consequently form a reward function in real-time as:
\begin{equation}
    r_{i,j}=w_{1} * \text{DM}_{i,j}+w_{2} * \left\|p^{ij}-p_t\right\|_2
\end{equation}
where $p^{ij} \in \mathbb{R}^{2\times 1}$ represents the pixel-wise position, $p_t=(t_x, t_y)$ is the target position of the tip calculated in Section II-A, $w_1$ and $w_2$ are weights for domain knowledge (\text{DM}) and moving cost. Finally, the target position is chosen from candidates whose r is larger than 95 percentile of overall $r$ value $Q_{0.95}\left(\boldsymbol{r}\right)$ with the minimum distance:
\begin{equation}
    \text{P}^{x,y}_\text{target} = \text{argmin}_{i,j}\left\|p^{ij}-p_t\right\|_2\text{.   }r.t\text{.  }r_{i,j} > Q_{0.95}\left(\boldsymbol{r}\right)
\end{equation}

In our target depth generator, we keep the target depth $d_\text{target}$ of the instrument within a certain interval $[8, 12]mm$ that is commonly used in MIS. 
Combining both 2D and 3D knowledge, the laparoscope can be well manipulated and place the main tool among a satisfactory region in the image.


\subsection{Laparoscopic Control}
\subsubsection{Misorientation Elimination}
\begin{figure}[thpb]
      \centering
      \includegraphics[width = 0.95\hsize]{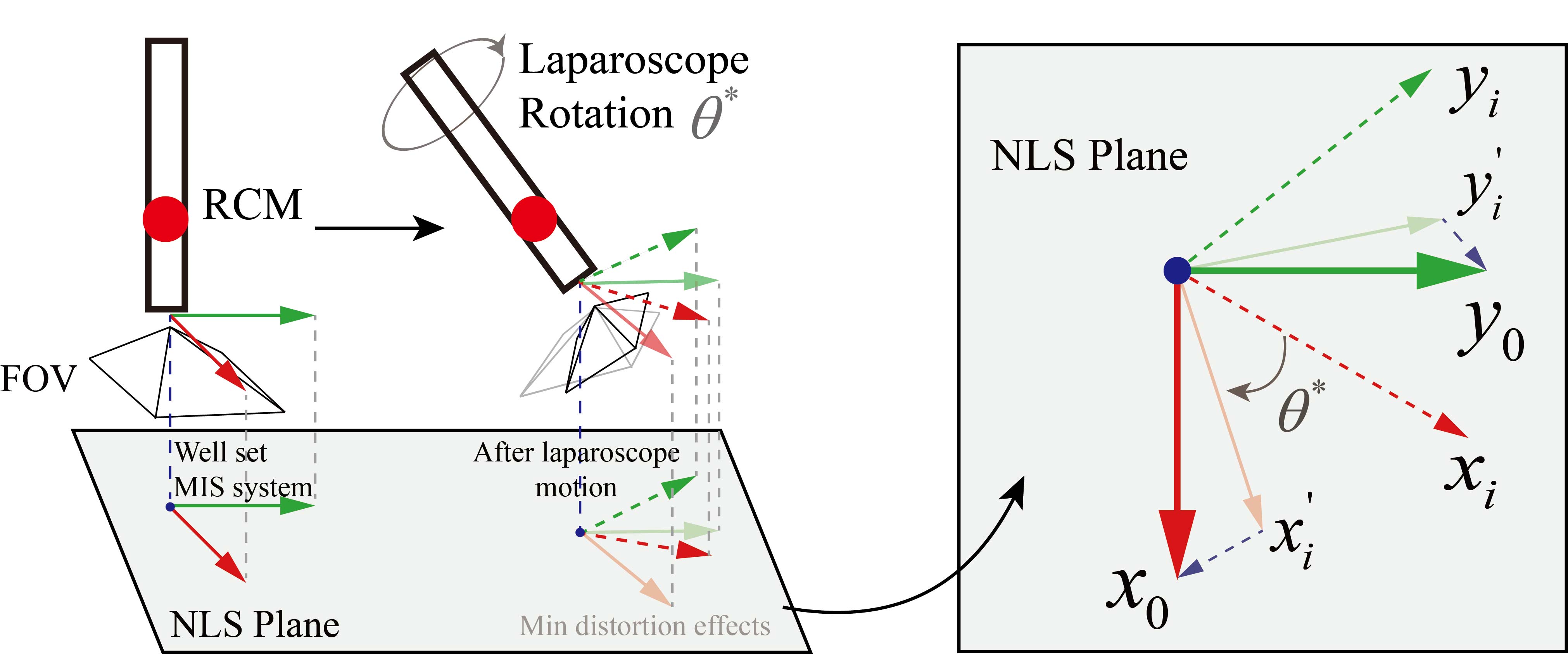}
      \caption{Illustration of misorientation effects. The thick axes $x_0,y_0$ represent orientation of a well-set MIS system, and the thin axes $x_i,y_i$ represent the orientation after the FOV control without any constraint. The thin axes $x_i^{\prime},y_i^{\prime}$ with light color represent the orientation adjusted using MRC.}
      \label{Misorientation explain and our solution}
   \end{figure}

To obtain an optimal FOV, our system should always provide visual feedback, which is consistent with the orientation of initial natural line-of-sight (NLS), for surgeons. Nevertheless, there always exits visual misorientation when the laparoscopic motion is constrained by RCM in MIS, which is shown in Fig. \ref{Misorientation explain and our solution}. 
To further reduce visual fatigue for surgeons, the laparoscopic orientation should be additionally controlled to minimize the difference of visual orientation in dynamic operations.

For our previous Intuitive Virtual Plane (IVP)-based method, it resolves this problem only when the tool is around the initial location, while the misorientation also occurs when the camera has large deviations.
To eliminate this ubiquitous issue, we introduce an evolved approach namely Minimize Rotation Constraint (MRC), in which we exploit the physical root of visual misorientation and use the affine mapping \cite{hartley2003multiple} to address this problem. The mapping can be expressed as:

\begin{equation}\left(p_0^{ij}, 1\right)^{\mathrm{T}} =\left(\begin{array}{ll}\textbf{A}_{\theta} & \textbf{t}_{\theta} \\ \textbf{0}_{1\times2} & 1\end{array}\right) \cdot
\left(p_{\theta}^{ij}, 1\right)^{\mathrm{T}}\label{distortion}\end{equation}
where $p_{0}^{ij}$ denotes the reference orientation at the initial time, $p_{\theta}^{ij}$ is the pose direction with a $\theta$ axial rotation angle of the laparoscope, $\textbf{A}_{\theta} \in \mathbb{R}^{2\times 2}$ and $\textbf{t}_{\theta}\in \mathbb{R}^{2\times 1}$ accordingly represent the distortion matrix and the 2D image displacement under axial rotation $\theta$.
For displacement $\textbf{t}_{\theta}$, it can be resolved by moving the laparoscope to put the instrument at the desired 2D position, which is mentioned in Section II-A.
In order to find an optimal $\theta$ to minimize the misorientation, we apply the following SVD composition to $\textbf{A}_{\theta}$:
\begin{equation}
\small
\begin{split}
\textbf{A}_{\theta} = UDV^{\mathrm{T}} &= \left(UV^{\mathrm{T}}\right) \left(VDV^{\mathrm{T}}\right)=\textbf{R}\left(\phi\right)\cdot \left(VDV^{\mathrm{T}}\right)
\end{split}
\end{equation}
where $\textbf{R}(\phi)$ depicts one misorientation component induced by the laparoscopic rotation $\theta$. To minimize this camera rotation-related misorientation, our target is to find a $\theta^{*}$ to minimize $|\phi|$, which can be calculated as: $\theta^{*}=\operatorname{argmin}_{\theta}|\phi|$. 


\subsubsection{Controller Design} 
By enhancing the controller in our previous work \cite{yang2019adaptive}, we refine the image Jacobian matrix ${\mathbf{J}}_{img}(p_t(t)) =\mathbf{J}_{p,v}\left(p_t, d_\text{tool}\right)$ by integrating our estimated scale-aware depth, which hence can improve control effect, where $\mathbf{J}_{p,v}\in \mathbb{R}^{2 \times 3}$ mapping from 2D image velocities to laparoscope velocities, is the first three columns of the image Jacobian matrix $\mathbf{J}_{p} \in \mathbb{R}^{2 \times 6}$.
Here a null space controller is proposed which adjusts the depth error $e_d=d_\text{tool} - d_\text{target}$ and misorientation $\theta^*$ simultaneously.
\begin{equation}\nonumber
\resizebox{1\hsize}{!}{
$\left[\begin{array}{l}
{^r\mathbf{v}_{r}(t)} \\
{^r\mathbf{\omega}_{r}(t)}
\end{array}\right]=
\left[\begin{array}{c}
{-\mathbf{K}_{r} \mathbf{e}_{r}(t)} \\
{-\mathbf{K}_{s} \mathbf{J}_{fov}^{+}(t) \mathbf{e}_{p}(t)-\underbrace{\left(\mathbf{I}_{4}-\mathbf{J}_{fov}^{+}(t) \mathbf{J}_{fov}(t)\right)}_{null \ space} \mathbf{J}_{de}^{+} \cdot [0\;0\; { }^c\mathbf{v}_{c}(t) | \;0\;0 \;{}^c\mathbf{\omega}_{c}(t)]^T}
\end{array}\right]$}
\end{equation}
where $\mathbf{K}_{r}$ is the positive definite gain matrix to control RCM error $e_r(t)$, $\mathbf{J}_{*}^{+}$ represents the pseudo inverse form of matrix $\mathbf{J}_{*}$, $\mathbf{J}_{de} =  [\mathbf{J}_{d}^T\; \mathbf{J}_{e}^T] ^T\in \mathbb{R}^{6 \times 4}$: $\mathbf{J}_{d}=\left[{e}_{3}| skew(^{r}\mathbf{t}_{c}(t))\right]$ with $z$ axis basis vector ${e}_{3}$ and translation vector $^{r}\mathbf{t}_{c}(t)$ from laparoscope to RCM frame, $\mathbf{J}_{e}=[0 | I_{3}]$ with the identity matrix $I_{3}$. $\mathbf{J}_{fov}=\mathbf{J}_{img}\cdot\mathbf{J}_{d}\in \mathbb{R}^{2 \times 4}$, ${{}^{c} \mathbf{v}_{c}(t) } = - k_d \cdot e_d$, and ${{}^{c} {\omega}_{c}(t)} = - k_{\theta} \theta^{*}$. You can refer to \cite{yang2019adaptive} for more detailed preliminary modeling process.

Then we convert the RCM velocity $\left[{^r\mathbf{v}_{r}(t)}, {^r\mathbf{\omega}_{r}(t)}\right]$ into the end-effector velocity $\left[{^b\mathbf{v}_{e}(t)}, {^b\mathbf{\omega}_{e}(t)}\right]$:
\begin{equation}
\resizebox{0.85\hsize}{!}{
$\left[\begin{array}{c}
^b\mathbf{v}_{e}(t) \\
^b\mathbf{\omega}_{e}(t)
\end{array}\right]=\left[\begin{array}{cc}
{ }^{b} \mathbf{R}_{r}(t) & \boldsymbol{0} \\
\boldsymbol{0} & { }^{b} \mathbf{R}_{r}(t)
\end{array}\right]\left[\begin{array}{cc}
\mathbf{I}_{3} & -skew\left(^r{\mathbf{t}}_{e}(t)\right) \\
\mathbf{0} & \mathbf{I}
\end{array}\right]\left[\begin{array}{c}
{^r\mathbf{v}_{r}(t)} \\
{^r\mathbf{\omega}_{r}(t)}
\end{array}\right]$}
\end{equation}
where $^{b} \mathbf{R}_{r}(t)$ denotes the rotation matrix from RCM frame to the UR5 base frame and $^r{\mathbf{t}}_{e}(t)$ is the translation vector from UR5 end-effector to RCM frame.

The asymptotic convergence of RCM error ${\mathbf e}_r$, tip position error ${\mathbf e}_{p}$, and depth error ${e}_{d}$ can be guaranteed by considering the Lyapunov-like quadratic function $V=\frac{1}{2}{\mathbf e}_{r}^{T}{\mathbf e}_{r}+\frac{1}{2}{\mathbf e}_{p}^{T}{\mathbf e}_{p}+\frac{1}{2}e_{d}^{2}$ such that $\Dot{V}=-{\mathbf e}_{p}^T(t){\mathbf{J}}_{fov}(t){\mathbf K}_s{\mathbf{J}}_{fov}^+(t){\mathbf e}_{p}(t) -{\mathbf e}_{r}^T(t) {\mathbf K}_{r} {\mathbf e}_{r}(t) -{k}_{d} {e}_{d}(t)^2< 0$ which proves the stability.

\section{EXPERIMENT RESULT}
\subsection{Experiment Setup}
The experiment platform shown in Fig. \ref{System Setup} consists of a UR5 robot, a Karl Storz laparoscope system, and a uterus phantom. 
The tool tip segmentation model is implemented in Pytorch, trained for 50 epochs using Adam solver with $\beta_1$=0.9 and $\beta_2$=0.999, the learning rate is set as $10^{-3}$, and the batch size is 2. The RoboDepth model is trained using Adam for 30 epochs with a batch size of 12. 
The inference results consisting of tip position and corresponding depth were send to the controller which was running on Robot Operating System (ROS) through UDP communication.
In our controller, the parameters were selected as follows: $\mathbf{K}_{s}$=diag(3e-3, 1, 1, 1), the RCM feedback gain $\mathbf{K}_{r}$=diag(0.5, 0.5), $k_\theta$=1, $k_d$=0.1.
We also added motion limits for robotic joints to compensate for occasional yet unexpected factors, e.g. depth estimation deviations, external disturbances, and etc., to further guarantee operational safety in practice.

\subsection{Performance of Tool Tip Segmentation}
To assess the tip segmentation performance, we adopted Intersection over Union (IoU), Precision, Recall and F1 score to comprehensively evaluate our enhanced segmentation model. 
In addition, we particularly proposed a metric named as Instrument Tip Position Accuracy (ITPA), in which it can be referred as a success only when the estimated tip position is located within the tool mask.

In our experiments, we used two sets of image data to quantitatively evaluate the performance of our tip segmentation model. Set 1 contains images collected based on the same phantom with that used in the training set, but with varying lighting conditions. As for Set 2, it was established by adopting a different surgical trainer as the background.
We totally leveraged 76 images in two sets, and the detailed results are accordingly listed in Table. \ref{table_segmentation}. 
\begin{table}[ht]
\caption{Testing results for tool tip segmentation. Unit: Percentage}
\begin{tabular}{|c||ccccc|}
\hline
Metrics & IoU & Dice & Recall & Precision & ITPA\\
\hline\hline
Set 1 & 84.8$\pm$6.5 & 91.7$\pm$3.9 & 89.7$\pm$7.7 & 94.5$\pm$5.5 & 100.0\\
\hline
Set 2 & 73.7$\pm$15.4 & 83.8$\pm$11.4 & 86.0$\pm$8.5 & 83.3$\pm$16.2 & 87.9\\
\hline
\end{tabular}
\label{table_segmentation}
\end{table}

It is worth noticing that our model achieved 100\% ITPA accuracy in Set 1, and obtained an average IoU of 84.8\%, as well as maintaining satisfactory performances regarding Dice, Recall, and Precision. In addition, it reveals our model can quickly learning the metal features of the tool tip, which shows average quantitative results of 87.9\% in ITPA accuracy and 73.7\% in IoU even when the background is variant from the training condition.
Regarding these outcomes, our model can provide a reliable tip segmentation and 2-D localization under different conditions, proving a solid foundation for the following tasks.


\begin{figure}[b]
      \centering
      \includegraphics[width = 0.95\hsize]{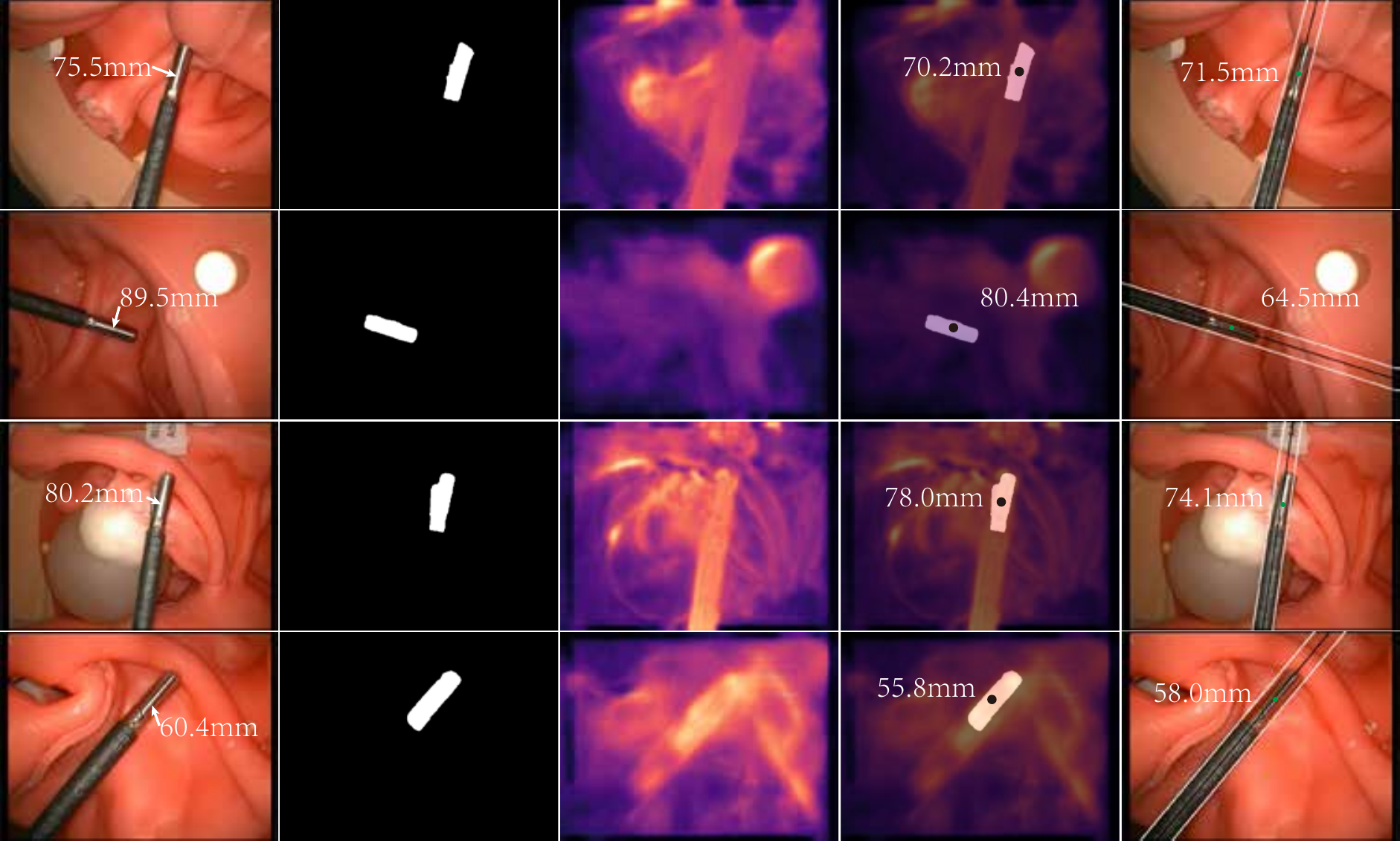}
      \caption{Qualitative results of the tool tip segmentation and depth estimation. Five columns are: 1. The original image with ground truth for tip depth; 2. Tip segmentation; 3. Dense scale-aware depth estimation by RoboDepth model; 4. Tip depth estimation combined the tool segmentation with the dense depth estimation; 5.Traditional method \cite{doignon2007degenerate} for depth estimation of tool tip (the black line is the reprojection of estimated cylinder axis).}
      \label{Depth_seg_results}
\end{figure}

\begin{table}[h]
\caption{Quantitative comparison of our method to the traditional tool pose estimation based method in each depth interval}
\label{Instrument tip depth estimation results}
\centering
\begin{tabular}{|c|l||c c | c|}
\hline
Model & Depth($mm$) & Abs Rel(\%) & RMSE($mm$) & Time\\
    \hline\hline
    \multirow{4}{*}{\textbf{Trad}} 
                          & Part 1: [4, 8] & 13.16$\pm$10.07 & 7.30$\pm$5.07 & \\
                          & Part 2: [8, 12] & 10.09$\pm$2.05 & 10.58$\pm$2.02 & \\
                          & Part 3: [12,16] & 18.33$\pm$6.16 & 28.3$\pm$12.58 & 0.70s\\
                          \cline{2-4}
                          & Overall: [4, 16] & 13.21$\pm$7.54 & 12.32$\pm$9.66 & \\
    \hline\hline
    \multirow{4}{*}{\textbf{\textcolor{red}{Ours}}} 
                          & Part 1: [4, 8] & 5.77$\pm$5.21 & 2.95$\pm$2.34 & \\
                          & Part 2: [8, 12] & 10.29$\pm$4.32 & 11.17$\pm$6.78 & \\
                          & Part 3: [12,16] & 9.11$\pm$7.53 & 12.67$\pm$9.71 & 0.02s\\
                          \cline{2-4}
                          & Overall: [4, 16] & 9.62$\pm$7.54 & 7.34$\pm$7.20 &\\
    \hline
\end{tabular}
\label{table_depth_estimation}
\end{table}

\subsection{Performance Assessment of Tool Tip Depth Estimation}
\begin{figure*}[h]
    \centering
    \includegraphics[width=0.99\textwidth]{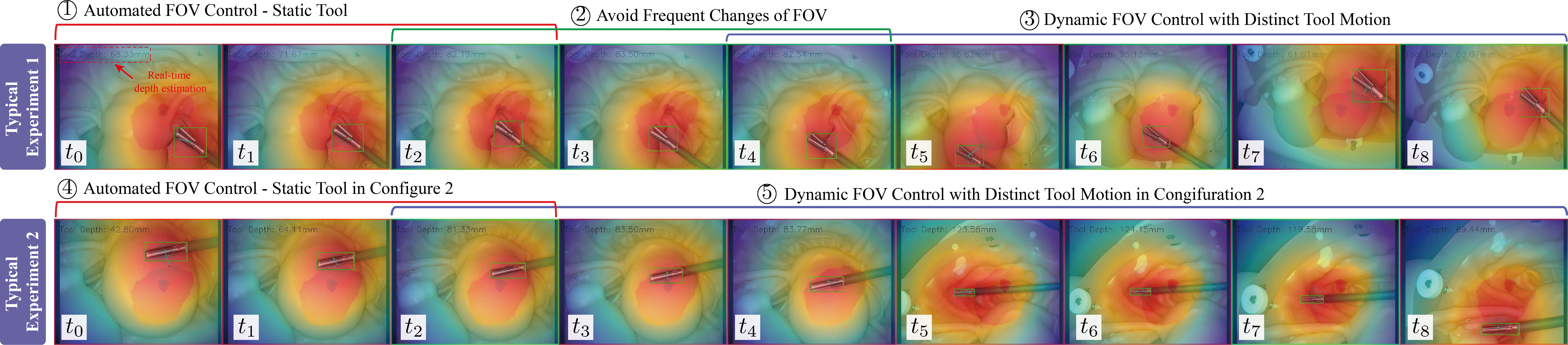}
    \caption{Typical results of automated FOV control based on a uterus trainer for hysterectomy.}
    \label{fig:experimental_snapshots}
    \vspace{-0.3cm}
\end{figure*}

To evaluate this unsupervised depth estimation, metrics including the Absolute Relative (Abs Rel) and root-mean-square error (RMSE) were adopted. 
We assessed the depth estimation performance by putting the tool tip within a [4, 16]$mm$ depth range, which is separated by every $4mm$.
Since all depths are scale-aware values, they can be directly compared with the measured ground truths.
Moreover, we also compared our results with those calculated based on the traditional cylinder instrument pose estimation \cite{doignon2007degenerate}, quantitative results are summarized in Table. \ref{table_depth_estimation} and qualitative results are shown in Fig. \ref{Depth_seg_results}. As can be seen that based on reliable tool tip segmentation, our RoboDepth model can hence recover its scale-aware depth in multiple conditions.

The results indicate that the inference time of our RoboDepth model is only 20 $ms$, which is nearly 35x faster than the traditional method.
Such superiority in computing speed gives a great advantage to our model in its practical implementation for online depth estimation. 
Besides, the overall Abs Rel of our RoboDepth model in depth estimation was reduced to 9.62\% with a comparison of 13.21\% obtained by the traditional method.
It is also worth noticing that even the largest Abs Rel for RoboDepth was 10.29\% within in [8, 12]$mm$, which is still quite precise enough for the laparoscope to adjust its depth and enable faster convergence than the constant depth-based method. 
\subsection{Elimination of Visual Misorientation}
\begin{figure}[thpb]
      \centering
      \includegraphics[width = 1.0\hsize]{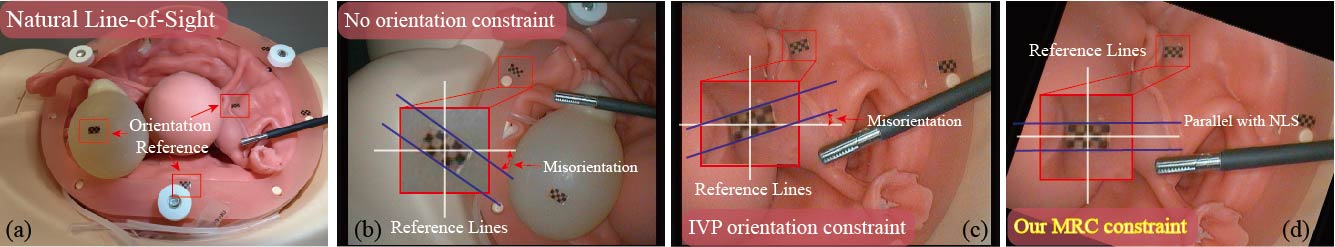}
      \caption{(a) Image captured in the natural of light direction. Snapshots of the control process: (b). without any constraint; (c). with IVP constraint; (d). with our proposed MRC constraint.}
      \label{Misorientation results}
   \end{figure}

To observe the improvement brought by our proposed MRC method clearly, the instrument was moved, w.r.t. the laparoscope, from an initial position to a farther location in a spiral fashion. 
In Fig. \ref{Misorientation results}, we showed the NLS image and relevant experimental snapshots, including the laparoscopic images without any constraint, with IVP constraint \cite{yang2019adaptive}, and with our MRC constraint. 
Several parallel-placed calibration boards were adopted as rotation references to reveal the orientation quality of visual feedback compared to the NLS. 

We can notice that the orientation without any constraint has the largest visual misorientation, which is around $30.0^{\circ}$ as is depicted in Fig. \ref{Misorientation results}(b). Although using additional constraints in IVP method, it still suffers a $18.0^{\circ}$ misorientation when the FOV has relatively large changes.
In contrast, the typical snapshot in Fig. \ref{Misorientation results}(c) suggests satisfactory visual feedback obtained using our MRC constraint, and it outputs almost the same orientation when comparing with the initial direction of NLS, which indicates the effectiveness of our approach for the correction of visual orientation.


\subsection{Results of Automated Optimal FOV Control}
\begin{figure}[thpb]
\centering
\includegraphics[width = 1.0\hsize]{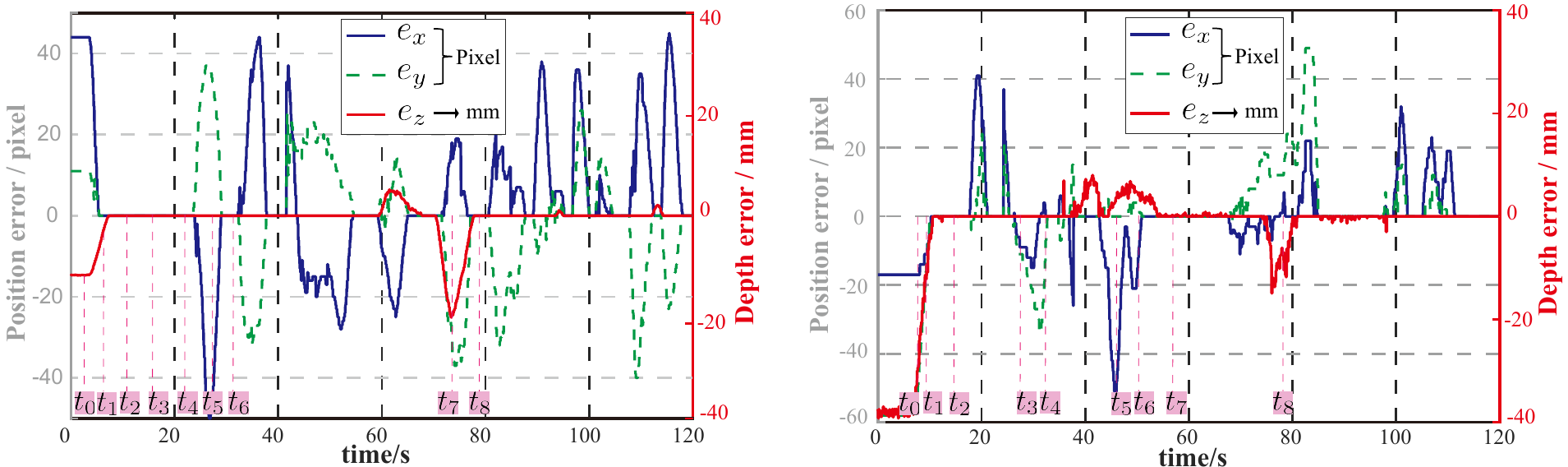}
\caption{Errors of automated FOV control in X-Y (image level) and Z (scale-aware depth) directions. Left: Experiment $\texttt{C}_1$; Right: Experiment $\texttt{C}_2$.}
\label{experiment_table}
\end{figure}

By integrating all function blocks together, the holistic framework was further validated on our platform (Fig. \ref{System Setup}(a)). Two typical experiments of automated FOV control denoted as $\texttt{C}_1$ and $\texttt{C}_2$ were shown in Fig. \ref{fig:experimental_snapshots}, and their corresponding control errors ($x$ and $y$ are in image level, $z$ is scale-aware depth) versus operating time were shown in Fig. \ref{experiment_table}.

As a starting point of optimal view control, the process between $t_0$ to $t_2$ in $\texttt{C}_1$ and $\texttt{C}_2$ shows the period of controlling the laparoscope to provide an optimal view for a static tool, and its 2-D image position error together with the scale aware depth error converge to zero asymptotically. 
Processes after $t_3$ demonstrate the optimal view control for a moving tool. 
As noticed in Fig. \ref{fig:experimental_snapshots}, when the tool is manipulated within a certain area between $t_2$ and $t_4$ in Experiment $\texttt{C}_1$, our domain knowledge-based control scheme can avoid frequent view changes, which satisfies the criteria proposed in Section II-B. 
Otherwise, when the tool moves to the new position which is away from the optimal view (e.g. from $t_5$ to $t_8$ in $\texttt{C}_1$, from $t_2$ to $t_8$ in $\texttt{C}_2$), our optimal view generator will output corresponding optimal viewpoints (red dot in the image) and calculate 3-D errors. By leveraging our null-space controller, the framework can autonomously move the laparoscope and reduce errors to zero.
It can be quantitatively seen from Fig. \ref{experiment_table} that, our controller can enable the scale-aware depth error within 20 $mm$ and the 2-D position error within 50 $pixels$.
All these results fully demonstrate the feasibility and reliability of our framework for automated FOV control.

\section{CONCLUSIONS AND FUTURE WORKS}

In this paper, we propose a novel data-driven framework for an automated laparoscopic optimal view control. A modified U-Net model and a novel unsupervised RoboDepth model are proposed to acquire an accurate 3-D perception of surgical tool tip. 
Besides, a new constraint with physical meaning named MRC is proposed to eliminate the visual misorientation problem. Compared with conventional control schemes, we also proposed an optimal view generator that can effectively extract control domain knowledge from large unlabeled surgical videos and perform FOV control in an expert-like manner.  

In the future, we will achieve more domain knowledge for camera control in more surgical scenarios, as well as proposing a generic framework to reliably automate laparoscopic motion under multi-instruments operation environments.

\bibliographystyle{IEEEtran}
\bibliography{ref}

\end{document}